\documentclass[]{fairmeta}

\usepackage{amsmath,amsfonts,bm}
\usepackage{xspace}
\usepackage{hyperref}
\usepackage{url}
\usepackage{subcaption}

\usepackage{colortbl}
\usepackage{xcolor}

\usepackage{graphicx}
\usepackage{algorithmic}
\usepackage{algorithm}
\usepackage{float}

\usepackage{booktabs}

\usepackage{enumitem}

\usepackage{amsthm}

\theoremstyle{plain}

\theoremstyle{definition}

\theoremstyle{remark}

\def\figref#1{figure~\ref{#1}}

\def\secref#1{section~\ref{#1}}

\def\eqref#1{equation~\ref{#1}}

\def\1{\bm{1}}

\DeclareMathAlphabet{\mathsfit}{\encodingdefault}{\sfdefault}{m}{sl}
\SetMathAlphabet{\mathsfit}{bold}{\encodingdefault}{\sfdefault}{bx}{n}

\usepackage[T1]{fontenc}

\usepackage[utf8]{inputenc}

\usepackage{microtype}

\IfFileExists{inconsolata.sty}{\usepackage{inconsolata}}{}

\usepackage{graphicx}
\usepackage[export]{adjustbox}
\usepackage{subcaption}
\usepackage{booktabs}
\usepackage{enumitem}
\usepackage{amsmath}
\usepackage{float}
\usepackage{xcolor}
\usepackage{hyperref}
\usepackage{tikz}
\usetikzlibrary{arrows.meta,positioning,shapes.geometric,fit,backgrounds}

\newcommand{\tabref}[1]{Table~\ref{#1}}
\renewcommand{\figref}[1]{Fig.~\ref{#1}}
\renewcommand{\secref}[1]{\S\ref{#1}}
\newcommand{\appref}[1]{Appendix~\ref{#1}}
\title{Representing Visual Evidence for Item Difficulty Prediction: Visual Textualization and Image-Native Modeling}

\author[1,*]{Han Chen}
\author[1,2,*]{Ming Li}
\author[2]{Hong Jiao}
\author[1]{Tianyi Zhou}

\contribution[*]{Co-first Author}

\renewcommand\affiliation[2][]{%
  \addtolist[#1]{#2}{\affiliationlist}{\affiliationformat}{\\}%
}

\affiliation[1]{Mohamed bin Zayed University of Artificial Intelligence}
\affiliation[2]{University of Maryland}

\abstract{Predicting item difficulty from content can provide an initial estimate for newly developed questions before sufficient student responses are available. Existing approaches typically represent the question stem and answer choices as text. When mathematics items contain visual components, a common pipeline first textualizes that evidence and then applies a text predictor. We ask: \emph{how should visual evidence be represented for item difficulty prediction?} We compare question text alone, \emph{visual textualization}, which expresses visual evidence in language, and \emph{image-native modeling}, which retains the original image. Using Eedi items with difficulty calibrated from student responses, we train large language models (LLMs) and vision-language models (VLMs) directly for difficulty regression. Both visual interfaces achieve the lowest point estimates, although the leading systems cannot be reliably ordered. Open-VLM textualization yields lower RMSE point estimates for all evaluated LLMs, while broader adaptation does so for all image-native VLMs. Test-time interventions show dependence on the paired full-item image, but do not isolate the additional visual component. The two visual interfaces also make partially complementary item-level errors and differ substantially in computational workflow. Thus, textualization should not be treated as the only practical interface: image-native modeling is a competitive alternative whose effectiveness depends on how the VLM is adapted.}

\date{\today}
\authoremails{\email{\{minglii, hjiao\}@umd.edu},
\email{\{han.chen, tianyi.zhou\}@mbzuai.ac.ae}}

\metadata[Project Page]{\url{https://github.com/MingLiiii/Visual_Item_Difficulty}\\}

\begin{document}
\maketitle

\section{Introduction}

Item difficulty is central to assessment design, item selection, and the sequencing of questions. Its psychometric calibration, however, requires observed student responses and is therefore unavailable for newly developed items before administration \citep{rasch1993probabilistic,hambleton1991fundamentals,demars2010item}. Predicting difficulty from item content can provide an initial estimate in this cold-start setting, supporting item development and preliminary use until response-based calibration becomes available \citep{alkhuzaey2021systematic,alkhuzaey2024text,peters2025text}. Content-based prediction thus complements rather than replaces empirical calibration.

\begin{figure*}[t]
    \centering
    \includegraphics[
        width=\textwidth,
    ]{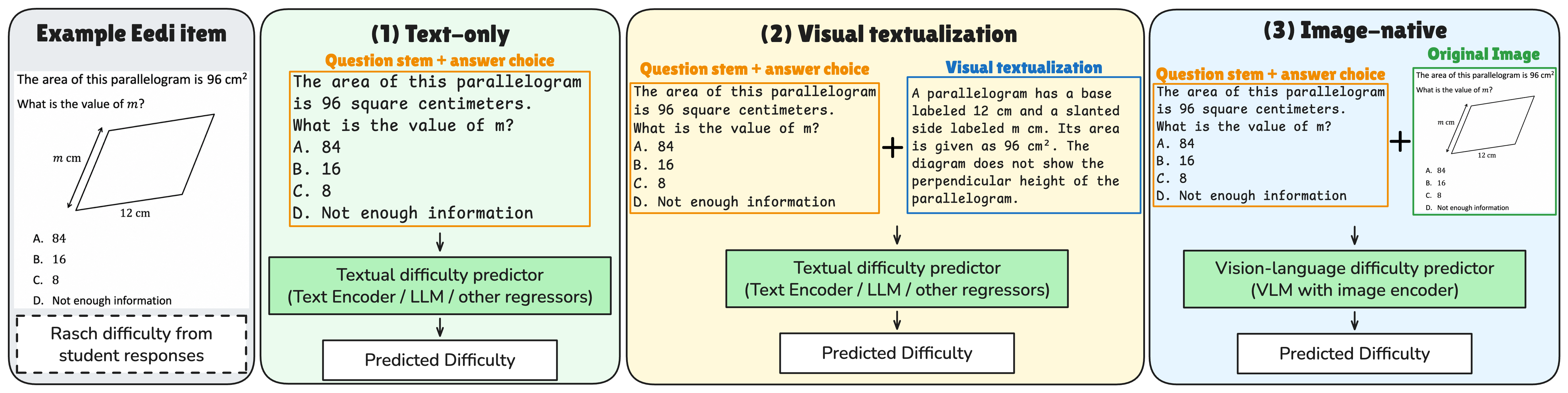}
    \caption{Three interfaces for content-based item difficulty prediction. All three receive the same question text $Q$, defined as the stem and answer choices of the specific question. The text-only route predicts from $Q$; visual textualization augments $Q$ with a fixed description $D$ of the additional visual component. The image-native modeling retains the original image $I$ alongside $Q$. 
    }
    \label{fig:modeling-paths}
    \vspace{-2mm}
\end{figure*}

\begin{figure*}[t]
\centering
\begin{subfigure}[t]{0.32\linewidth}
\centering
\includegraphics[width=\linewidth]{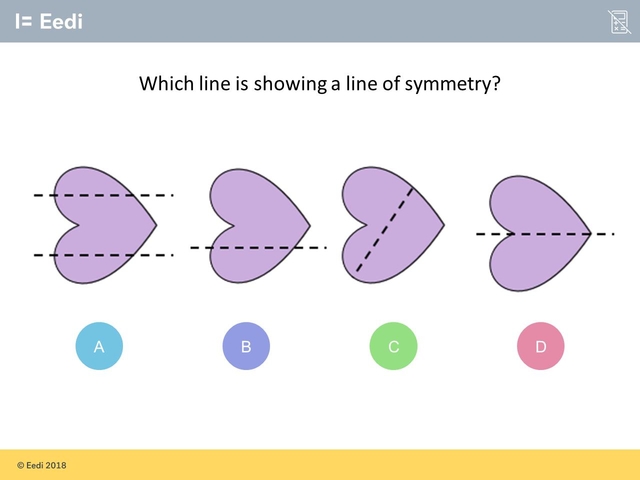}
\caption{Visual answer options}
\label{fig:item-examples-symmetry}
\end{subfigure}
\hfill
\begin{subfigure}[t]{0.32\linewidth}
\centering
\includegraphics[width=\linewidth]{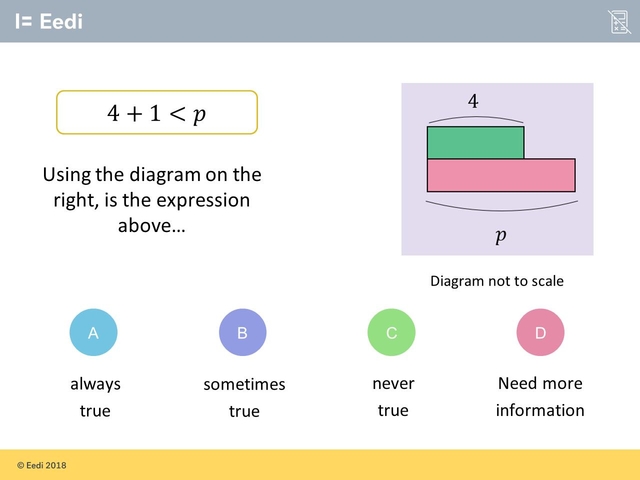}
\caption{Not-to-scale relation}
\label{fig:item-examples-scale}
\end{subfigure}
\hfill
\begin{subfigure}[t]{0.32\linewidth}
\centering
\includegraphics[width=\linewidth]{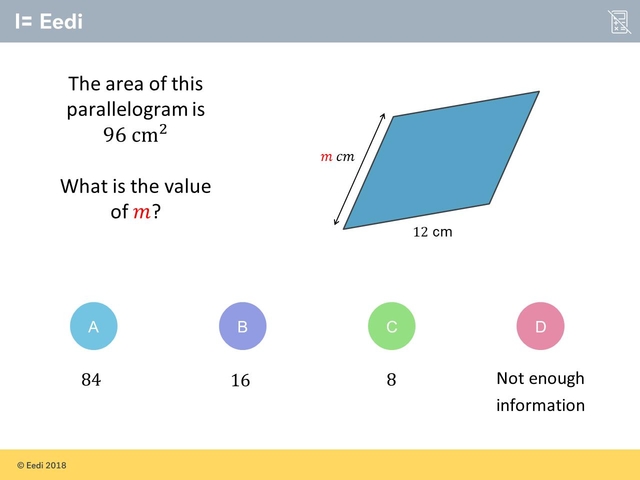}
\caption{Slanted side vs.\ height}
\label{fig:item-examples-geometry}
\end{subfigure}
\caption{Training-split examples spanning three recurring visual-representation challenges, selected before any model-error analysis. In (a), the answer options are visual configurations. In (b), a textualizer must preserve the distinction between depicted and asserted relations in a not-to-scale diagram. In (c), the marked segment is a slanted side rather than a perpendicular height. \appref{app:error-cases} provides a systematic item-type analysis.}
\label{fig:item-examples}
\end{figure*}

Mathematics assessment items often include diagrams or other visual elements that affect how students interpret and solve them
\citep{sweller1988cognitive,sweller2011cognitive,noroozi2022scrutiny, chen2021geoqa,lu2024mathvista}. We call items whose solution-relevant content extends beyond the text of the stem and answer choices \emph{visually grounded assessment items}. \figref{fig:item-examples} illustrates three cases from Eedi data used in our experiments: the additional evidence lies in a visual answer configuration, a geometric relation, or the distinction between a marked side and a perpendicular height. Expressing such evidence in language requires a textualizer to decide which relations to state, how to linearize spatial structure, and how to describe ambiguous or not-to-scale depictions.

Modern vision-language models (VLMs)
\citep{bai2025qwen25vl,bai2025qwen3,zhu2025internvl3,
wang2025internvl3,steiner2024paligemma2}
enable two practical approaches. In \textbf{visual textualization}, a VLM converts the visual component into a fixed description for a text predictor; unlike ordinary captioning, the description records problem-relevant notation and spatial relations. In \textbf{image-native modeling}, the original image remains available to the final VLM predictor. To our knowledge, these interfaces have not been systematically compared for response-calibrated item difficulty prediction, nor have image-native VLMs been directly adapted to this continuous target.

This leads to our central question: \emph{how should visual evidence be represented for item difficulty prediction?} We compare two practical workflows, \textbf{visual textualization} and \textbf{image-native modeling}, against a text-only setting. \figref{fig:modeling-paths} summarizes the three modeling paths. All systems receive the same question stem and answer-choice text; the two visual workflows differ in whether additional evidence is supplied as generated language or retained in the full-item image. Holding the prediction target, item split, and evaluation protocol fixed supports a common evaluation of these practical interfaces, while their inputs and final predictor families remain distinct. We evaluate the approaches on Eedi mathematics items from the NeurIPS 2020 Education Challenge \citep{wang2020diagnostic}, using Rasch difficulty parameters derived from student responses as prediction targets \citep{rasch1993probabilistic,hambleton1991fundamentals,demars2010item}.

Crucially, our main LLM and VLM predictors undergo supervised task adaptation to these response-derived targets rather than being evaluated only through prompted inference. Our experiments span four text encoders, three vision encoders, five LLMs, and ten VLMs; the LLMs and VLMs range from 2B to 8B parameters. Across these model classes, we evaluate task-adapted regression, frozen-feature regression, scalar generation, and late fusion. The strongest systems using item text alone, visual textualization, and
image-native modeling reach 0.517, 0.506, and 0.497 RMSE, respectively.
Visual textualization yields lower RMSE point estimates for all five
matched text models, while broader adaptation does so for all ten
image-native VLMs. Test-time interventions show dependence on the
paired full-item image, but do not isolate the additional visual
component. The two visual strategies also divide item-level wins almost evenly and have substantially different workflow costs. Together, these findings show that visual-evidence representation is a consequential modeling choice with no uniformly dominant solution and should be evaluated explicitly in item difficulty prediction.

Our contributions are:
\begin{itemize}[leftmargin=*]
 \vspace{-2mm}
 \item We provide a systematic comparison of three practical interfaces for item difficulty prediction, item text alone, visual textualization, and image-native modeling, under a shared response-calibrated target and evaluation protocol.
 \vspace{-2mm}
 \item We evaluate 22 text and vision models up to 8B parameters on Eedi mathematics items, including supervised adaptation of ten VLMs for continuous regression alongside frozen-feature, scalar-generation, and late-fusion baselines.
  \vspace{-2mm}
\item We find consistent point-estimate reductions from textualization
across five text models and from broader adaptation across ten VLMs.
Image-native prediction depends on the paired full-item image, and the
two visual interfaces differ in cost and item-level errors.
\end{itemize}

\section{Related Work}
Content-based difficulty prediction has progressed from handcrafted linguistic features to pretrained representations and language models \citep{perkins1995predicting,loukina2016textual,xue2020predicting,li2025item}. Most work assumes text input, whereas multimodal mathematics requires interpreting diagrams, notation, and spatial relations \citep{chen2021geoqa,lu2024mathvista}. Recent analyses further show that multimodal performance does not by itself establish visual dependence \citep{liu2025role,wang2025benchmarking}. We therefore compare practical visual interfaces for response-calibrated difficulty regression rather than problem solving. \textbf{A detailed related work section can be found in \appref{sec:related_work}.}

\section{Task and Data}
\label{sec:task-data}

\subsection{Dataset}
\label{sec:dataset}

We use the Eedi dataset released for the NeurIPS 2020 Education Challenge \citep{wang2020diagnostic}. The challenge is based on diagnostic mathematics questions answered by students on the Eedi platform. Unlike many item difficulty prediction benchmarks that assume clean text inputs or use extracted textual item features \citep{alkhuzaey2021systematic,alkhuzaey2024text,peters2025text,li2025item}, the item content in this dataset is distributed as original question images. Each image contains the question stem, answer choices, and, when applicable, mathematical notation, figures, diagrams, tables, and spatial layout. This makes the dataset a natural testbed for studying difficulty prediction for visually grounded assessment items.

We focus on the question set used in Tasks 3 and 4 of the challenge, which contains student response records for a shared pool of mathematics items \citep{wang2020diagnostic}. Since our goal is to predict item-level difficulty from item content, we construct an item-level dataset rather than a student-response prediction dataset. We remove items marked as deleted in the released metadata and items with fewer than 200 observed student responses, so that the downstream difficulty estimates are based on a minimum amount of response evidence. After filtering, the resulting dataset contains 725 items. We use a fixed item-level split with 580 items for training and model selection and 145 held-out items for testing. All models are evaluated on the same held-out test items.

\subsection{Difficulty Labels}
\label{sec:difficulty-labels}

The original challenge does not provide item difficulty labels for our task. We therefore use item parameters estimated from student correctness records with a one-parameter logistic item response model, also known as the Rasch model \citep{rasch1993probabilistic,hambleton1991fundamentals,demars2010item}. For each student--item interaction, the observed answer is represented by a binary correctness indicator $y_{ui} \in \{0,1\}$. The model defines the probability that student $u$ answers item $i$ correctly as
\begin{equation}
\begin{aligned}
P(y_{ui}=1 \mid \theta_u, \beta_i)
&= \sigma(\theta_u - \beta_i) \\
&= \frac{1}{1 + \exp[-(\theta_u - \beta_i)]},
\end{aligned}
\end{equation}
where $\theta_u$ is the latent ability of student $u$ and $\beta_i$ is the difficulty of item $i$. The item parameter $\beta_i$ is used as the prediction target, with larger values corresponding to more difficult items. The released response records are used to count the observations supporting each parameter; after excluding items with fewer than 200 responses, the retained estimates range from approximately $-2.91$ to $1.96$. Retained items have a median of 2,155 observed responses (range 203--2,966). Their reported difficulty standard errors have mean 0.061, median 0.054, and range 0.043--0.162.

This label construction is important for the interpretation of the task. The target is not a manually annotated quality score and is not supplied by the Eedi metadata. It is a psychometric estimate derived from how students answered each item, following the item response theory tradition of calibrating item parameters from examinee response data \citep{rasch1993probabilistic,hambleton1991fundamentals,demars2010item}. We predict the item difficulty parameter itself, not its standard error. A sensitivity analysis excludes items with the largest reported standard errors; the ordering of the three displayed systems is unchanged (\appref{app:label-uncertainty}).

\subsection{Item Representations}
\label{sec:item-representations}

The Eedi release provides each item as a single question image rather than as separate text and visual fields \citep{wang2020diagnostic}. From this source, we define three representations: the original image $I$, the question text $Q$, and a generated description $D$ of any distinct visual component.

\paragraph{Original image.}
The original image $I$ contains the complete rendered item, including the stem, answer options, mathematical notation, and layout; \figref{fig:item-examples} shows representative examples. Of the 725 items, 401 (55.3\%) contain an additional visual component: 325 of 580 training items (56.0\%) and 76 of 145 test items (52.4\%). For the remaining 324 items, $D$ is empty.

\paragraph{Question text ($Q$).}
We extract the question stem and answer options from $I$ using the
OpenAI GPT-5.5 API alias \texttt{gpt-5.5}
\citep{openai2026gpt55systemcard}, accessed June 2026, then apply a second model-based checking pass and manual inspection. We therefore treat $Q$ as a \emph{manually verified, image-derived transcription}, not as raw OCR or ground-truth text. It excludes the separate visual description $D$, so a $Q$-only predictor receives neither $I$ nor $D$. The necessary exception is the 18 items whose answer choices are images: $Q$ includes short manually verified descriptions of those choices so that the options are not blank; four such items occur in the test set. Exact extraction procedures appear in \appref{app:parsing-prompts}, and a matched raw-OCR control appears in \appref{app:ocr-control}.

\paragraph{Visual description ($D$).}
For each of the 401 items with an additional visual component, we generate a fixed textual description of that component. Unlike $Q$, these descriptions are not manually verified or treated as item text; they are experimental visual textualizations used to make additional evidence available to a text predictor. We compare descriptions generated by GPT-5.5, denoted $D_{\mathrm{GPT}}$, and Qwen2.5-VL-7B, denoted $D_{\mathrm{VL7B}}$~\citep{openai2026gpt55systemcard,bai2025qwen25vl}. Text predictors receive either $Q$ or $Q{+}D$, whereas image-native VLMs retain $I$ in the prediction interface, optionally together with $Q$ or $Q{+}D$.

\section{Representing Visual Evidence}
\label{sec:modeling-paradigms}

We organize systems by the representation available to the final predictor. The question-text setting uses $Q$ alone, visual textualization uses $Q{+}D$, and image-native modeling uses $I{+}Q$. In all three main settings, the predictor is trained on the same training split. The comparison therefore concerns not whether a pretrained model can produce a zero-shot difficulty judgment, but how each representation supports supervised difficulty prediction.

\subsection{Problem Formulation}
\label{sec:problem-formulation}

Using the representations defined in \secref{sec:item-representations}, the three main inputs for item $i$ are
\begin{equation}
    x_i \in \{Q_i,\,[Q_i;D_i],\,(I_i,Q_i)\}.
\end{equation}
Given one of these inputs, a predictor $f_\phi$ estimates
\begin{equation}
    \hat{\beta}_i = f_\phi(x_i).
\end{equation}
The parameterization of $f_\phi$ depends on the modeling setting. It may include all model weights under full fine-tuning, task-specific adapters and a regression readout, or an external regressor fitted to frozen representations. Regression-based systems minimize mean squared error:
\begin{equation}
    \mathcal{L}_{\mathrm{reg}} = \frac{1}{N}\sum_{i=1}^{N}(\hat{\beta}_i - \beta_i)^2.
\end{equation}
Architecture-specific objectives, readouts, and optimization details appear in \appref{app:auxiliary-methods}, \appref{app:text-implementation}, and \appref{app:vlm-implementation}.

\subsection{Prediction from Question Text}
\label{sec:transcription-only-methods}

The question-text setting predicts
\begin{equation}
\hat{\beta}_i=f_{\mathrm{text}}(Q_i).
\end{equation}
Because $Q_i$ is manually verified after extraction, this setting is deliberately stronger than raw OCR. It measures what can be predicted from the stem and answer-choice text without access to either the original image or a generated description of the additional visual component. We evaluate this interface across multiple text-model families and fitting strategies rather than tie it to a single architecture or adaptation method.

\subsection{Prediction via Visual Textualization}
\label{sec:textualization-methods}

Visual textualization separates representation construction from difficulty prediction. A fixed VLM textualizer $T_m$ first generates $D_i$ from the image, after which a supervised text model predicts from $Q_i{+}D_i$:
\begin{equation}
D_i=T_m(I_i), \qquad
\hat{\beta}_i=f_{\mathrm{text}}([Q_i;D_i]).
\end{equation}
The descriptions are generated once and are not optimized using difficulty labels. We evaluate both $D_{\mathrm{GPT}}$ and $D_{\mathrm{VL7B}}$ \citep{openai2026gpt55systemcard, bai2025qwen25vl}; for the 324 items without an additional visual component, $D_i$ is empty and the input reduces to $Q_i$. We use \emph{visual textualization} rather than captioning because $D_i$ records problem-relevant notation and spatial or geometric relations. Although the final predictor consumes only text, the pipeline is vision-informed because $D_i$ is generated from $I_i$. We compare $Q{+}D$ with $Q$ rather than use $D$ alone, since a $D$-only condition would remove the stem and answer choices as well as change access to visual evidence.

\subsection{Image-Native VLM Prediction}
\label{sec:image-native-vlm-methods}

Image-native modeling retains the original image as an input to the supervised final predictor:
\begin{equation}
\hat{\beta}_i=f_{\mathrm{VLM}}(I_i,Q_i).
\end{equation}
We instantiate this route with Qwen-VL
\citep{bai2025qwen25vl,bai2025qwen3}, InternVL
\citep{zhu2025internvl3,wang2025internvl3}, and PaliGemma
\citep{steiner2024paligemma2} families. The term \emph{image-native} refers to this direct access to $I_i$ rather than to an image-only model: the image preserves the original notation, layout, and graphical relations, while $Q_i$ provides a stable, checked rendering of the stem and answer choices across VLM families. We use $I+Q$ as the canonical image-native input so that every VLM receives both the original visual artifact and the same curated linguistic channel.

For image-native regression, we vary pooling and the scope of parameter-efficient adaptation across language and vision components. These experiments train the VLM for the difficulty task rather than query it only at inference time; complete target modules, family-specific exceptions, and optimization details appear in \appref{app:vlm-implementation}.

\paragraph{Auxiliary paradigms.}
We additionally evaluate scalar generation, frozen text, vision, and VLM representations, and late fusion as secondary design-space checks. Their objectives and readouts appear in \appref{app:auxiliary-methods}; they are not part of the unified three-seed representation comparison.

\begin{table*}[!t]
\centering
\tiny
\setlength{\tabcolsep}{3.3pt}
\renewcommand{\arraystretch}{0.94}
\begin{adjustbox}{width=0.94\linewidth,center}
\begin{tabular}{@{}lllp{1.35cm}p{2.15cm}rr@{}}
\toprule
Paradigm & Model & Input & Adaptation & Readout & RMSE $\downarrow$ & $\rho\uparrow$ \\
\midrule
\emph{Text encoder} & BERT-base & $Q+D$ & Full FT & Reg.\ head & 0.5585 & 0.7330 \\
& RoBERTa-base & $Q+D$ & Full FT & Reg.\ head & 0.6050 & 0.6664 \\
& DeBERTa-v3-base & $Q+D$ & Full FT & Reg.\ head & 0.5410 & 0.7528 \\
& ModernBERT-base & $Q+D$ & Full FT & Reg.\ head & 0.5908 & 0.6885 \\
\midrule
\emph{Text LLM} & Qwen2.5-3B & $Q+D$ & Attn LoRA & Reg.\ head & 0.5266 & 0.7553 \\
& Llama-3.2-3B & $Q+D$ & Attn LoRA & Reg.\ head & 0.5335 & 0.7672 \\
& Qwen2.5-7B & $Q+D$ & Attn LoRA & Reg.\ head & 0.5169 & 0.7791 \\
& Qwen3-8B & $Q+D$ & Attn LoRA & Reg.\ head & 0.5210 & 0.7749 \\
& Llama-3.1-8B & $Q+D$ & Attn LoRA & Reg.\ head & \underline{0.5059} & \textbf{0.7939} \\
\cmidrule(l){2-7}
& Qwen2.5-3B & $Q+D$ & Frozen & XGBoost & 0.5541 & 0.7436 \\
& Llama-3.2-3B & $Q+D$ & Frozen & XGBoost & 0.5292 & 0.7828 \\
& Qwen2.5-7B & $Q+D$ & Frozen & XGBoost & 0.5454 & 0.7723 \\
& Qwen3-8B & $Q+D$ & Frozen & Ridge & 0.5383 & 0.7627 \\
& Llama-3.1-8B & $Q+D$ & Frozen & XGBoost & 0.5437 & 0.7736 \\
\cmidrule(l){2-7}
& Qwen2.5-3B & $Q+D$ & LoRA SFT & Token generation & 0.9096 & 0.0977 \\
& Llama-3.2-3B & $Q+D$ & LoRA SFT & Token generation & 0.8698 & 0.0844 \\
& Qwen2.5-7B & $Q+D$ & LoRA SFT & Token generation & 0.8987 & 0.0331 \\
& Qwen3-8B & $Q+D$ & LoRA SFT & Token generation & 0.7646 & 0.3385 \\
& Llama-3.1-8B & $Q+D$ & LoRA SFT & Token generation & 0.7901 & 0.3443 \\
\midrule
\emph{Vision encoder} & CLIP ViT-L/14 & $I$ & Frozen & Shallow reg. & 0.5993 & 0.6729 \\
& SigLIP-so400m & $I$ & Frozen & Shallow reg. & 0.5704 & 0.6990 \\
& DINOv2-large & $I$ & Frozen & Shallow reg. & 0.5565 & 0.7212 \\
\midrule
\emph{VLM} & Qwen2.5-VL-3B & $I+Q$ & Attn+MLP & Reg.\ head & 0.5280 & 0.7586 \\
& Qwen2.5-VL-7B & $I+Q$ & Attn+MLP & Reg.\ head & \textbf{0.4966} & \underline{0.7853} \\
& Qwen3-VL-4B & $I+Q$ & Attn+MLP & Reg.\ head & 0.5306 & 0.7547 \\
& Qwen3-VL-8B & $I+Q$ & Attn+MLP & Reg.\ head & 0.5277 & 0.7631 \\
& PaliGemma2-3B & $I+Q$ & Attn+MLP & Reg.\ head & 0.5611 & 0.7206 \\
& InternVL3-2B & $I+Q$ & Attn+MLP & Reg.\ head & 0.5452 & 0.7384 \\
& InternVL2.5-4B & $I+Q$ & Attn+MLP & Reg.\ head & 0.5214 & 0.7590 \\
& InternVL3-8B & $I+Q$ & Attn+MLP & Reg.\ head & 0.5500 & 0.7268 \\
& InternVL3.5-4B & $I+Q$ & Attn+MLP & Reg.\ head & 0.5111 & 0.7685 \\
& InternVL3.5-8B & $I+Q$ & Attn+MLP & Reg.\ head & 0.6137 & 0.6513 \\
\cmidrule(l){2-7}
& Qwen2.5-VL-3B & $I+Q$ & Frozen & LightGBM & 0.5599 & 0.7243 \\
& Qwen2.5-VL-7B & $I+Q$ & Frozen & XGBoost & 0.5450 & 0.7418 \\
& PaliGemma2-3B & $I+Q$ & Frozen & LightGBM & 0.5753 & 0.7186 \\
& InternVL2.5-4B & $I+Q$ & Frozen & XGBoost & 0.5536 & 0.7300 \\
& InternVL3-8B & $I+Q$ & Frozen & XGBoost & 0.5833 & 0.6916 \\
& InternVL3.5-8B & $I+Q$ & Frozen & Ridge & 0.5804 & 0.7237 \\
\cmidrule(l){2-7}
& Qwen2.5-VL-3B & $I+Q$ & LoRA SFT & Token generation & 0.8688 & 0.3533 \\
& Qwen2.5-VL-7B & $I+Q$ & LoRA SFT & Token generation & 0.6030 & 0.7028 \\
& Qwen3-VL-4B & $I+Q$ & LoRA SFT & Token generation & 0.6455 & 0.6133 \\
& Qwen3-VL-8B & $I+Q$ & LoRA SFT & Token generation & 0.6010 & 0.6619 \\
& PaliGemma2-3B & $I+Q$ & LoRA SFT & Token generation & 0.6700 & 0.5654 \\
\midrule
\emph{Late fusion} & BGE + SigLIP & $Q+D \oplus I$ & Frozen & LightGBM & 0.5791 & 0.6915 \\
& BGE + Qwen2.5-VL-3B & $Q+D \oplus (I+Q)$ & Frozen & XGBoost & 0.5534 & 0.7357 \\
& BGE + DINOv2 + Qwen2.5-VL-3B & $Q+D \oplus I \oplus (I+Q)$ & Frozen & XGBoost & 0.5518 & 0.7454 \\
\bottomrule
\end{tabular}
\end{adjustbox}
\caption{Performance across representation and modeling paradigms. The table spans text and vision encoders, LLMs, VLMs, and late fusion; subdivisions within the LLM and VLM blocks separate task-adapted regression, frozen-feature regression, and scalar generation. Unless otherwise noted, $D$ denotes the visual description generated by
Qwen2.5-VL-7B, $D_{\mathrm{VL7B}}$; description-source and VLM-adaptation comparisons appear in \tabref{tab:textualization-scale} and \tabref{tab:vlm-mlp}. Bold and underlining mark the best and second-best result for each metric.}
\vspace{-4mm}
\label{tab:route-comparison}
\end{table*}

\section{Experiments and Results}
\label{sec:experiments-results}

\subsection{Experimental Setup}
\label{sec:experimental-setup}

The unified comparison evaluates five text LLMs and ten VLMs with at most 8B parameters on a fixed split of 580 training and 145 test items. Training recipes are selected using only the training split and then fixed before final three-seed evaluation. Exact checkpoints, model-selection procedures, and training configurations appear in \appref{app:data-eval} to \appref{app:vlm-implementation}.

We report RMSE as the primary metric and Spearman correlation as a secondary metric, summarized across seeds. For paired comparisons, we ensemble predictions across seeds and use a paired item-level bootstrap to obtain 95\% confidence intervals for differences in RMSE; full statistical details appear in \appref{app:statistics}.

Matched comparisons vary representation, adaptation scope, or the
paired full-item image while holding the relevant model or predictor
fixed. Complete results appear in the appendix.

\subsection{Performance Across Representations}
\label{sec:route-results}

\tabref{tab:route-comparison} provides a broad map of the evaluated design space, spanning text and vision encoders, LLMs, VLMs, multiple adaptation and readout strategies, and late fusion. Within the unified three-seed task-adapted regression results, the interface-level leaders are Llama-3.1-8B with attention+MLP adaptation on $Q$, Llama-3.1-8B with attention-only adaptation on $Q{+}D_{\mathrm{VL7B}}$, and Qwen2.5-VL-7B with attention+MLP adaptation on $I{+}Q$. They reach 0.517, 0.506, and 0.497 RMSE, respectively, giving the two visual interfaces the lowest point estimates in this comparison. Because these interface leaders use different models or adaptation scopes, they summarize the strongest configuration for each representation rather than a matched estimate of the representation effect; matched comparisons follow in later subsections. Across the broader table, performance is not monotonic in model size, underscoring the need to compare representations across multiple model families and fitting strategies.

The point estimates rank image-native modeling first, visual textualization second, and question text alone third. However, all three paired bootstrap intervals include zero (\appref{app:statistics}), so the test set does not support a reliable ordering among these interface-level leaders. We therefore treat them as a competitive group and use the matched analyses below to study the two visual strategies.

\begin{table}[t]
\centering
\small
\setlength{\tabcolsep}{2.8pt}
\begin{adjustbox}{width=0.43\linewidth,center}
\begin{tabular}{@{}l|c|cc|c@{}}
\toprule
Model & $Q\downarrow$ & $Q{+}D_{\mathrm{G}}\downarrow$ & $Q{+}D_{\mathrm{V}}\downarrow$ & $\Delta_{\mathrm{V}}\downarrow$ \\
\midrule
Qwen2.5-3B & 0.5367 & 0.5343 & 0.5266 & $-0.0094$ \\
Llama-3.2-3B & 0.5615 & 0.5608 & 0.5335 & $-0.0270$ \\
Qwen2.5-7B & 0.5226 & 0.5258 & 0.5169 & $-0.0048$ \\
Qwen3-8B & 0.5380 & \textbf{0.5165} & 0.5210 & $-0.0147$ \\
Llama-3.1-8B & \textbf{0.5210} & 0.5298 & \textbf{0.5059} & $-0.0146$ \\
\bottomrule
\end{tabular}
\end{adjustbox}
\caption{Matched visual-textualization results with the text model and training recipe fixed. $D_{\mathrm{G}}$ and $D_{\mathrm{V}}$ denote GPT-5.5 and Qwen2.5-VL-7B descriptions. Input columns report three-seed mean RMSE; $\Delta_{\mathrm{V}}$ compares seed-ensemble predictions for $Q{+}D_{\mathrm{V}}$ and $Q$. Full confidence intervals appear in \tabref{tab:app-textualization-ci}; bold marks the lowest mean in each input column.}
\label{tab:textualization-scale}
\vspace{-4mm}
\end{table}

\paragraph{Encoder baselines.}
Conventional encoders capture substantial difficulty signal but do not match the strongest task-adapted generative backbones as shown above. The best fully fine-tuned text encoder reaches 0.541 RMSE, compared with 0.506 for the best adapted text LLM on the same $Q{+}D$ interface. The best frozen vision encoder reaches 0.557, whereas the strongest task-adapted image-native VLM reaches 0.497. These comparisons show that neither compact textual encoding nor generic visual features alone generate leading results.

\paragraph{Fitting and output interfaces.}
The fitting and output interface matters within both LLM and VLM blocks. The best frozen text and VLM regressors reach 0.529 and 0.545 RMSE, respectively, while the best scalar-generation systems reach 0.765 and 0.601. Late fusion of frozen representations reaches 0.552 and therefore does not obtain the performance of task-adapted multimodal regression. The best RMSE and Spearman results also come from different systems: Qwen2.5-VL-7B attains the lowest RMSE, while Llama-3.1-8B with visual textualization attains the highest rank correlation. Absolute calibration and item ordering are therefore related but distinct aspects of performance.

\subsection{Matched Visual Textualization Lowers RMSE Point Estimates}
\label{sec:textualization-results}

To isolate the effect of visual textualization, we hold the downstream text model and attention-only LoRA recipe fixed and vary only its input: $Q$, $Q{+}D_{\mathrm{G}}$, or $Q{+}D_{\mathrm{V}}$. Here, $D_{\mathrm{G}}$ and $D_{\mathrm{V}}$ are descriptions generated by GPT-5.5 and Qwen2.5-VL-7B, respectively. \tabref{tab:textualization-scale} reports this matched comparison for all five text models.

Adding $D_{\mathrm{V}}$ lowers RMSE relative to $Q$ for all five models, with seed-ensemble differences from $-0.0048$ to $-0.0270$. The confidence interval excludes zero for Llama-3.2-3B but includes zero for the other four models (\tabref{tab:app-textualization-ci}). Thus, the direction is consistent across the evaluated models, but the evidence for any individual improvement is generally imprecise. The Qwen2.5-VL-7B description also yields a lower mean RMSE than the GPT-5.5 description for four of five downstream models.
Although the final predictor is text-based, the $Q{+}D$ conditions are vision-informed because a VLM has inspected the image before prediction. The consistent point-estimate reductions suggest that verbalized visual evidence can help without changing the downstream model. Differences between $D_{\mathrm{GPT}}$ and $D_{\mathrm{VL7B}}$ further indicate that the textualizer is part of the modeling choice rather than interchangeable preprocessing.

\begin{table}[t]
\centering
\scriptsize
\setlength{\tabcolsep}{2.2pt}
\begin{adjustbox}{width=0.48\linewidth,center}
\begin{tabular}{@{}l|rrr|r@{}}
\toprule
VLM & Attn. $\downarrow$ & +MLP $\downarrow$ & $\rho_{\mathrm{MLP}}\uparrow$ & $\Delta_{\mathrm{M}}\downarrow$ \\
\midrule
Qwen2.5-VL-3B & 0.5762 & 0.5280 & 0.7586 & $-0.0510$ \\
Qwen2.5-VL-7B & 0.5233 & \textbf{0.4966} & 0.7853 & $-0.0260$ \\
Qwen3-VL-4B & 0.5461 & 0.5306 & 0.7547 & $-0.0171$ \\
Qwen3-VL-8B & 0.5379 & 0.5277 & 0.7631 & $-0.0116$ \\
PaliGemma2-3B & 0.5857 & 0.5611 & 0.7206 & $-0.0266$ \\
InternVL3-2B & 0.5765 & 0.5452 & 0.7384 & $-0.0304$ \\
InternVL2.5-4B & 0.5562 & 0.5214 & 0.7590 & $-0.0341$ \\
InternVL3-8B & 0.5806 & 0.5500 & 0.7268 & $-0.0321$ \\
InternVL3.5-4B & 0.5408 & 0.5111 & 0.7685 & $-0.0293$ \\
InternVL3.5-8B & 0.6354 & 0.6137 & 0.6513 & $-0.0336$ \\
\bottomrule
\end{tabular}
\end{adjustbox}
\caption{Matched image-native adaptation results for $I{+}Q$. Attn.\ and +MLP report three-seed mean RMSE for attention-only and attention+MLP LoRA; $\rho_{\mathrm{MLP}}$ is Spearman correlation for +MLP. $\Delta_{\mathrm{M}}$ is the seed-ensemble RMSE difference for +MLP minus attention-only, so negative values favor broader adaptation. Full confidence intervals appear in \tabref{tab:app-vlm-mlp-ci}.}
\label{tab:vlm-mlp}
\vspace{-4mm}
\end{table}

\subsection{Adaptation Breadth in Image-Native Prediction}
\label{sec:image-native-vlm-results}

We next ask whether image-native regression is limited by adapting too narrow a portion of the VLM. \tabref{tab:vlm-mlp} holds the $I+Q$ input and regression interface fixed while extending LoRA from attention projections to both attention and MLP projections. All ten models have negative $\Delta_{\mathrm{M}}$, meaning that attention+MLP adaptation lowers seed-ensemble RMSE; the mean difference is $-0.029$, and six confidence intervals exclude zero. Shared data and related model families mean that these systems are not independent replicates; the descriptive sign test appears in \appref{app:statistics}.

The analogous change is much smaller for the matched Llama-3.1-8B $Q$-only control: it lowers the three-seed mean RMSE by 0.0043 and seed-ensemble RMSE by 0.0066, with a confidence interval of $[-0.0314,0.0214]$. This contrast suggests that the VLM gains are not merely a generic consequence of adding MLP LoRA targets. Instead, they are consistent with an adaptation bottleneck in image-native regression, where attention-only updates may be insufficient to reshape multimodal representations for a continuous psychometric target.

\subsection{Test-Time Dependence on the Paired Image}
\label{sec:image-intervention}

An image-native interface does not guarantee that the trained predictor uses its image, because $Q$ already provides the question and
answer-choice text; related work has similarly questioned whether
multimodal mathematical reasoning systems genuinely depend on their visual inputs \citep{liu2025role,wang2025benchmarking}. \tabref{tab:image-intervention} summarizes this intervention for the strongest attention+MLP configuration.

\begin{table}[t]
\centering
\small
\begin{adjustbox}{width=0.48\linewidth,center}
\begin{tabular}{lcc}
\toprule
Test image & RMSE $\downarrow$ & Spearman $\uparrow$ \\
\midrule
Original & $\mathbf{0.4966\pm0.0062}$ & $\mathbf{0.7853\pm0.0079}$ \\
Blank & $0.9447\pm0.0475$ & $0.6788\pm0.0069$ \\
Shuffled & $0.9643\pm0.0059$ & $0.1118\pm0.0065$ \\
\bottomrule
\end{tabular}
\end{adjustbox}
\caption{Test-time image interventions for the Qwen2.5-VL-7B $I{+}Q$ attention+MLP predictor (three-seed mean $\pm$ SD). Text input and trained parameters are fixed.}
\label{tab:image-intervention}
\vspace{-4mm}
\end{table}

Using a blank image increases RMSE from 0.497 to 0.945, while pairing each item with another item's image increases RMSE to 0.964 and reduces Spearman correlation from 0.785 to 0.112. These results show that the predictor depends on the paired full-item
image rather than ignoring the image stream. Because the interventions
also alter image-rendered text and layout, they do not isolate dependence on the additional visual component. Results for attention-only $I+Q$
and image-only prediction appear in
\appref{app:image-intervention}.

\section{Representation Trade-offs}
\label{sec:further-discussion}

\subsection{Choosing a Visual-Evidence Interface}
\label{sec:discussion-text-only}

The aggregate results do not establish an absolute ordering between visual textualization and image-native modeling, but the two interfaces impose different constraints. 

Visual textualization produces an inspectable language representation that can be cached, audited, edited, and reused across predictors. Its usefulness depends on which relations the textualizer selects and how it expresses them, as reflected by the differences between the two description sources. Image-native modeling retains the source image and allows its use to be shaped by the difficulty objective, but must learn this mapping from labeled items and process the image during every training and inference run. The practical choice therefore depends on whether reusable and inspectable language outweighs direct, task-conditioned access to notation, layout, and graphical relations.

The two leading visual strategies also make complementary item-level errors. Image-native modeling has lower absolute error on 74 of the 145 test items, while visual textualization performs better on 71; their residual correlation is 0.872. A fixed, untrained average reaches 0.478 RMSE, better than both component point estimates, although its paired intervals relative to them include zero. We therefore treat this result as evidence of complementarity rather than a reliably superior ensemble. Detailed subgroup, difficulty-stratified, and qualitative analyses appear in \appref{app:complementarity} and \appref{app:error-cases}.

Excluding the one-time generation of $D_{\mathrm{VL7B}}$, visual-textualization training and evaluation average 191 seconds,
compared with 1,602 seconds for image-native modeling. These
workflow-specific timings reflect different models, batch sizes, and
epoch counts. Textualization permits descriptions to be cached and
reused, whereas image-native modeling repeatedly processes images.
Further details appear in \appref{app:workflow-cost}.

\subsection{Training Implications for Image-Native Prediction}
\label{sec:discussion-adaptation}

The image-native results depend not only on access to the image but also on how the VLM is adapted. Extending LoRA from attention projections to the language backbone's MLP projections lowers the RMSE point estimate for all ten VLMs, with six paired intervals excluding zero. A plausible explanation is that the regression head reads hidden states after both attention and MLP transformations, so attention-only updates may leave an important part of the task-specific mapping fixed. The matched text control also covers one Llama model, so the evidence does not establish that the benefit is specific to multimodal architectures.

The output objective creates a second training choice. Rasch difficulty is continuous rather than a linguistic label \citep{rasch1993probabilistic,hambleton1991fundamentals,demars2010item}. Scalar generation performs substantially worse than regression under the evaluated recipes, making the continuous readout another consequential part of the image-native interface.

\section{Conclusion}
\label{sec:conclusion}

We studied how visual evidence should enter response-calibrated item
difficulty prediction by comparing question text, visual
textualization, and image-native modeling across a broad collection of
text and vision models. The strongest systems using either visual
interface achieve lower RMSE point estimates than the strongest
question-text system, although paired intervals do not reliably order
the three interface leaders. Matched analyses provide a more specific
picture: Qwen2.5-VL-7B textualizations lower RMSE point estimates across
all five downstream text models, while broader language-side adaptation
does so across all ten image-native VLMs. Test-time interventions further
show that the image-native predictor depends on the paired full-item
image, without establishing that its gain comes specifically from the
additional visual component.

\section*{Limitations}
\label{sec:limitations}

The primary limitation is the number and scope of calibrated items. Our experiments use 725 English-language mathematics items from a single assessment source, with 580 items for training and 145 for testing. Human-response-calibrated difficulty labels are inherently difficult to scale because each item must accumulate sufficient learner interactions before its difficulty can be estimated. In this dataset, each retained item is supported by at least 200 responses, with a median of 2,155 responses per item. The benchmark therefore contains substantial response evidence despite its modest number of item-level labels. Learning from a limited set of calibrated items is not only a constraint of this study but also an important practical setting for difficulty prediction, where collecting labels for additional items can require responses from many students.

\section*{Ethical Considerations}

This study uses the publicly released Eedi dataset and does not collect new participant data. We work with item content and de-identified response records provided by the benchmark rather than information intended to identify individual students. Content-based difficulty estimates should be used only as provisional support for item development and cold-start decisions; they should not replace empirical calibration from student responses or expert review. Prediction errors could otherwise lead to inappropriate item sequencing or assessment decisions, and performance may not transfer to other curricula, languages, populations, or accessibility contexts. Any deployment should therefore monitor subgroup performance where appropriate, retain human oversight, and recalibrate items using responses from the intended learner population.

\bibliographystyle{assets/plainnat}
\bibliography{main}

\clearpage
\tableofcontents
\newpage
\beginappendix

\section{Extended Related Work}
\label{sec:related_work}

\subsection{Content-Based Item Difficulty Prediction}

Automatic item difficulty prediction estimates item difficulty from item
content before sufficient student responses are available for empirical
calibration. Early work relied on handcrafted linguistic and
psycholinguistic features
\citep{perkins1995predicting,loukina2016textual,hsu2018automated,
yaneva2019predicting}, while later approaches adopted pretrained
representations, transfer learning, large language models, and interpretable
features extracted from reasoning traces
\citep{xue2020predicting,he2021automatically,mccarthy2021jump,
li2025item,feng2025reasoning,razavi2026estimating,wang2026cognitive}. The BEA 2024 shared task
further benchmarked statistical, machine-learning, and LLM-based systems
\citep{yaneva2024findings,tack2024itec,rogoz2024unibucllm,
duenas2024upn,veeramani2024large,fulari2024utilizing}; broader reviews
summarize the progression of the field
\citep{alkhuzaey2021systematic,alkhuzaey2024text,peters2025text}.

Most prior work, however, treats the item representation as given and
primarily operates on question text or text-derived features. This assumption
is restrictive for visually grounded mathematics items, where diagrams,
notation, spatial layout, and graphical relations may contain
solution-relevant information. We therefore study the representation of item
content itself as a modeling choice.

\subsection{Language Models for Psychometric Prediction}

Recent work has increasingly connected language models with psychometric
properties estimated from observed student responses. Fine-tuned LMs and LLMs
have been used for direct difficulty or item-parameter prediction
\citep{li2025item,han2025leveraging}, while other approaches estimate
difficulty through proficiency-conditioned prompting or simulated student
responses
\citep{park2024large,benedetto2024using,li2025can}. Related studies examine
whether LLM-generated responses exhibit plausible psychometric behavior and
emphasize validity constraints in student simulation
\citep{liu2025leveraging,sauberli2025llms,yuan2026towards}, while recent
results show that more
demanding item properties such as discrimination remain challenging
\citep{chen2026llms}.

Our target is likewise response-calibrated: we predict Rasch item difficulty
estimated from student responses
\citep{rasch1993probabilistic,hambleton1991fundamentals,demars2010item}.
Rather than changing the psychometric target or simulating examinees, we ask
how multimodal item content should be represented to a supervised predictor.

\subsection{Representing Visual Evidence in Multimodal Assessment Items}

Visual mathematics problems often require information that is not fully
recoverable from plain text. Benchmarks such as GeoQA and MathVista require
models to jointly interpret language with diagrams, geometric structure, and
other visual evidence
\citep{chen2021geoqa,lu2024mathvista}. More recent work has shown that strong
multimodal performance does not necessarily imply genuine dependence on the
visual modality, motivating explicit tests of image dependence and benchmarks
constructed around visually necessary evidence
\citep{liu2025role,wang2025benchmarking}. This distinction is particularly
relevant to assessment items, where a rendered figure may encode relations
that cannot be reconstructed from the stem and answer choices alone.

Visual evidence can enter a predictor through several interfaces. Generic vision encoders such as CLIP, SigLIP, and DINOv2 provide image representations learned from large-scale visual or image--text data \citep{radford2021learning,zhai2023sigmoid,oquab2023dinov2}. Modern vision-language models instead jointly process image and text, including Qwen2.5-VL \citep{bai2025qwen25vl}, Qwen3-VL \citep{bai2025qwen3}, InternVL2.5 \citep{chen2024expanding}, InternVL3 \citep{zhu2025internvl3}, InternVL3.5 \citep{wang2025internvl3}, and PaliGemma 2 \citep{steiner2024paligemma2}. These models make direct image-native prediction practical while retaining access to linguistic context.

An alternative is to convert the visual component into language before
prediction. We use the term \emph{visual textualization} for this interface:
a VLM verbalizes problem-relevant visual information such as labels,
dimensions, geometric relations, and spatial configurations, after which a
text model predicts difficulty from the augmented input. This differs from
ordinary OCR, whose primary objective is to recover visible characters rather
than encode graphical semantics. In our experiments, GPT-5.5
\citep{openai2026gpt55systemcard} and Qwen2.5-VL
\citep{bai2025qwen25vl} serve as textualizers.

Visual textualization and image-native modeling impose different
representational bottlenecks. Textualization produces a fixed, inspectable,
and reusable linguistic representation, but information omitted or
misexpressed by the textualizer is unavailable downstream. Image-native
modeling retains the original rendered item and allows the downstream
difficulty objective to determine how visual information is used, while
requiring repeated multimodal processing. Prior item-difficulty work has
largely varied prediction models while assuming textual input, whereas
multimodal mathematical reasoning work primarily evaluates problem solving
rather than response-calibrated psychometric prediction. To our knowledge,
these visual interfaces have not been systematically compared for item
difficulty prediction under a shared target, split, and evaluation protocol.

\section{Implementation Details}
\label{app:implementation-details}

\subsection{Data Split and Model Selection}
\label{app:data-eval}

All main experiments use the same fixed split of 580 training items and 145 test items.

The unified model scope contains five text LLMs and ten VLMs with at most 8B parameters. The text models are Qwen2.5-3B/7B \citep{qwen2024qwen25},
Llama-3.2-3B and Llama-3.1-8B \citep{grattafiori2024llama},
and Qwen3-8B \citep{yang2025qwen3}.
The VLMs are Qwen2.5-VL-3B/7B \citep{bai2025qwen25vl},
Qwen3-VL-4B/8B \citep{bai2025qwen3},
PaliGemma2-3B \citep{steiner2024paligemma2},
InternVL2.5-4B \citep{chen2024expanding},
InternVL3-2B/8B \citep{zhu2025internvl3},
and InternVL3.5-4B/8B \citep{wang2025internvl3}.

We partition the training items into five folds of 116 items, stratified jointly by difficulty quintile and the presence of an additional visual component. Hyperparameters are selected only from these folds. Using representative text and multimodal models, we select one family-level recipe and apply it to every model and input route in that family. We consider LoRA rank, learning rate, training duration, and a Huber-loss alternative; Huber loss does not improve the representative models. \tabref{tab:app-recipes} lists the final recipes.

Cross-validation selects the family-level training recipes before final training on all 580 items. The main comparison reports the strongest systems under each interface, while the complete tables document performance across all evaluated text models and VLMs. The matched $Q$ versus $Q+D$ analysis fixes each text model and recipe, and the attention versus attention+MLP analysis pairs two adaptation scopes for every VLM.

\begin{table*}[t]
\centering
\small
\begin{adjustbox}{width=0.52\linewidth,center}
\begin{tabular}{lcccc}
\toprule
Family & LoRA rank & Learning rate & Loss & Epochs \\
\midrule
Qwen text & 8 & $2\mathrm{e}{-4}$ & MSE & 7 \\
Llama text & 16 & $1\mathrm{e}{-4}$ & MSE & 6 \\
Qwen-VL & 8 & $5\mathrm{e}{-5}$ & MSE & 8 \\
InternVL 8B & 8 & $5\mathrm{e}{-5}$ & MSE & 5 \\
InternVL 2--4B & 8 & $5\mathrm{e}{-5}$ & MSE & 10 \\
PaliGemma2 & 8 & $5\mathrm{e}{-5}$ & MSE & 6 \\
\bottomrule
\end{tabular}
\end{adjustbox}
\caption{Frozen family-level recipes. The 10-epoch setting for small InternVL models follows training-side learning curves that continued to improve after epoch 5.}
\label{tab:app-recipes}
\end{table*}

For final evaluation, each configuration is trained on all 580 training items for the fixed epoch count in \tabref{tab:app-recipes}, with no validation split or early stopping. We run seeds 17, 42, and 2026.

\subsection{Item Parsing and Visual Textualization}
\label{app:parsing-prompts}

The initial parser uses the OpenAI GPT-5.5 API alias \texttt{gpt-5.5}, accessed in June 2026, and operates in two passes. The first pass extracts the question and identifies any additional visual component; the second checks the extraction against the same source image. The exact prompts are given below.

The extraction system prompt is:
\begin{quote}\small
You are a math question parser. Given an image of a multiple-choice math question from the Eedi platform, extract its content as a JSON object. Return only fields \texttt{question}, \texttt{choices} with keys A--D, \texttt{has\_figure}, and, only when applicable, \texttt{figure\_description}. Preserve mathematical symbols exactly and retain references to visual elements verbatim. If a choice is an image, provide a brief bracketed description and never leave it blank. Set \texttt{has\_figure} true for any problem-relevant diagram, shape, graph, number line, table, flowchart, or image-valued choice, excluding the answer-letter bubbles and Eedi header. Describe all figures concisely but completely, including labels, dimensions, arrows, shading, and spatial relations.
\end{quote}

The verification system prompt is:
\begin{quote}\small
You are a meticulous math question verifier. Compare the image against every field in the draft JSON. Correct missing or garbled mathematical symbols, truncated text, the \texttt{has\_figure} decision, and incomplete or inaccurate figure descriptions. Image-valued answer choices must receive brief bracketed descriptions. Return the corrected object with the same schema, no extra fields, and no explanation outside the JSON.
\end{quote}

The verification pass receives the same image and the complete draft extraction. We then manually inspect $Q$ against the source image. Difficulty values and response outcomes are not part of the parsing or inspection interface, preventing label-guided edits. The inspection targets the fidelity of the stem and choices; generated descriptions remain model outputs rather than human annotations. We did not retain an edit-level audit trail and therefore cannot quantify a manual correction rate.

For $D_{\mathrm{VL7B}}$, Qwen2.5-VL-7B receives the following fixed instruction for the 401 items with an additional visual component:
\begin{quote}\small
This image is from a math assessment item. In 1--3 sentences, describe the figure/diagram only: the visual elements (shapes, graphs, axes, geometry, labels, numbers) that a student would need to read to answer. Be concise and factual. Do not solve the question and do not restate the question text.
\end{quote}
Generation is limited to 128 new tokens. For the other 324 items, $D_{\mathrm{VL7B}}$ is empty.

\subsection{Text Models and Visual Textualization}
\label{app:text-implementation}

Text inputs are tokenized to a maximum length of 512. Training uses batch size 4 and gradient accumulation 4. The final hidden state is mean-pooled over non-padding tokens and passed to a regression head consisting of layer normalization, dropout, a linear projection to 256 dimensions, GELU, and a scalar output layer. The final text recipe applies LoRA to $q$, $k$, $v$, and output attention projections. We additionally evaluate attention+MLP LoRA on Llama-3.1-8B by including gate, up, and down projections.

For prediction through visual textualization, descriptions are available for the 401 items with an additional visual component. For all other items, $D$ is empty and $Q+D$ equals $Q$. We denote the GPT-5.5 descriptions by $D_{\mathrm{GPT}}$ and descriptions generated by Qwen2.5-VL-7B by $D_{\mathrm{VL7B}}$.

We do not treat $D$ alone as a matched information route. The description prompt deliberately excludes the question text, and $D$ is empty for 324 items. A $D$-only system would therefore conflate access to visual evidence with removal of the stem and answer choices. Our estimand is the incremental value of a visual interface conditional on the common curated transcription $Q$.

\subsection{Image-Native VLMs}
\label{app:vlm-implementation}

Image-native denotes the $I+Q$ representation, in which the original image remains available to the final predictor; it does not denote image-only input. VLM training uses batch size 1 and gradient accumulation 4. Images are processed with each model family's native processor. Qwen-VL and PaliGemma use their packaged image preprocessing; InternVL uses a $448\times448$ image transform and the model's image-context tokens. The pooled representation is taken from the language backbone's final hidden state.

The final image-native adaptation targets the language backbone's attention and MLP projections. In Qwen-VL and InternVL, these targets leave the visual tower frozen. PaliGemma2 uses shared projection names, so its visual attention projections are included in both sides of the matched attention-only versus attention+MLP comparison; the added MLP targets remain language-side. LoRA uses $\alpha=2r$, dropout 0.05, and no bias. We compare this configuration against:
\begin{itemize}[leftmargin=*]
\item \textbf{attention-only LoRA}, which targets $q$, $k$, $v$, and output projections;
\item \textbf{text-token pooling}, which excludes image placeholder and boundary tokens before pooling; and
\item \textbf{vision-encoder LoRA}, which additionally targets visual attention projections identified from their full module paths.
\end{itemize}

Across all neural models, optimization uses AdamW with weight decay 0.01, cosine decay, 10\% warmup, gradient clipping at 1.0, and bfloat16 arithmetic. The regression head and LoRA parameters are optimized jointly.

\subsection{Auxiliary Modeling Paradigms}
\label{app:auxiliary-methods}

\paragraph{Scalar generation.}
As an auxiliary output-interface comparison, we fine-tune text LLMs and VLMs to generate a standardized difficulty value as a JSON string rather than predict it with a regression head. For
\begin{equation}
z_i=\frac{\beta_i-\mu_{\mathrm{train}}}{\sigma_{\mathrm{train}}},
\end{equation}
the token-level objective is
\begin{equation}
    \mathcal{L}_{\mathrm{gen}}
    =-\sum_t \log p_\theta(s_{i,t}\mid s_{i,<t},x_i).
\end{equation}
At evaluation time, the generated value is parsed and transformed back to the Rasch scale. This baseline tests whether a standard language-generation interface is suitable for numeric psychometric prediction.

\paragraph{Frozen representations and late fusion.}
Frozen-feature baselines do not optimize the neural regression objective end to end. They extract fixed text, vision, or VLM representations and fit an external regressor selected by cross-validation on the training split. We evaluate frozen vision encoders and VLM representations as well as late fusion \citep{radford2021learning,zhai2023sigmoid,oquab2023dinov2}. Late fusion concatenates frozen text, vision, and/or VLM representations before fitting a shallow regressor. These comparisons test whether visual access or feature concatenation alone can match task-adapted image-native prediction. Representative results appear in \appref{app:auxiliary-results}.

\subsection{Representative Workflow Cost}
\label{app:workflow-cost}

We compare wall-clock time for the leading visual-textualization and image-native systems on the same hardware. The duration includes model loading, final fitting on 580 items, and prediction on 145 test items, but excludes the one-time generation of $D_{\mathrm{VL7B}}$.

\tabref{tab:app-workflow-cost} reports mean job time and mean time per epoch for the two representative workflows.

\begin{table*}[t]
\centering
\small
\begin{adjustbox}{width=0.58\linewidth,center}
\begin{tabular}{lrrr}
\toprule
Representation & Epochs & Mean job time (s) & Mean time/epoch (s) \\
\midrule
Visual textualization ($Q+D$) & 6 & 191.3 & 31.9 \\
Image-native ($I+Q$) & 8 & 1602.0 & 200.2 \\
\bottomrule
\end{tabular}
\end{adjustbox}
\caption{Representative workflow cost for the leading Llama-3.1-8B visual-textualization system and Qwen2.5-VL-7B image-native system. Times include model loading, final training, and test prediction, but
exclude the one-time generation of $D_{\mathrm{VL7B}}$. The
training-and-evaluation job ratio is $8.4\times$, and the
epoch-normalized ratio is $6.3\times$. Because the models, batch sizes, and epoch counts differ, these measurements characterize the evaluated workflows rather than intrinsic architecture efficiency.}
\label{tab:app-workflow-cost}
\end{table*}

The cached representation also changes how often images are processed. Visual textualization processes each relevant image once to generate $D$; subsequent regression operates entirely on text. Image-native training processes the images in every epoch and again at inference. This comparison does not assign a universal cost to either strategy, but it shows why the upstream textualization cost can be amortized when descriptions are reused.

\subsection{Uncertainty Estimates}
\label{app:statistics}

For each system, we report the mean and population standard deviation of test RMSE and Spearman correlation across the three seeds; compact main-text ablation tables omit some standard deviations, which are supplied below. For a paired comparison, we first average the three predictions for each test item within each system. We then resample the 145 paired items with replacement 10,000 times and recompute the difference in RMSE. The 2.5th and 97.5th percentiles form the reported confidence interval. This procedure preserves the item-level pairing and avoids treating the three training seeds as independent test sets. Intervals are not corrected for multiple comparisons and condition on
the fixed train--test split and selected training recipes; they do not
capture uncertainty due to alternative item splits,
hyperparameter-selection procedures, or dataset sampling.

For the three interface-level leaders, visual textualization changes seed-ensemble RMSE by $-0.0080$ relative to question text alone (95\% CI $[-0.0356,0.0189]$), and image-native modeling changes it by $-0.0148$ ($[-0.0519,0.0198]$).

The direct image-native minus visual-textualization difference is $-0.0068$ ($[-0.0520,0.0381]$). These seed-ensemble differences need not equal differences between the run-average RMSE values in the main table because the former average predictions before computing RMSE.

\tabref{tab:app-textualization-ci} gives the matched textualization difference and confidence interval for each downstream text model. \tabref{tab:app-vlm-mlp-ci} gives the corresponding comparison between attention+MLP and attention-only adaptation for each image-native VLM.

\begin{table}[t]
\centering
\small
\begin{adjustbox}{width=0.48\linewidth,center}
\begin{tabular}{@{}lrr@{}}
\toprule
Text model & $\Delta_{\mathrm{V}}\downarrow$ & 95\% CI \\
\midrule
Qwen2.5-3B & $-0.0094$ & $[-0.0302, 0.0111]$ \\
Llama-3.2-3B & $-0.0270$ & $[-0.0473,-0.0072]$ \\
Qwen2.5-7B & $-0.0048$ & $[-0.0260, 0.0155]$ \\
Qwen3-8B & $-0.0147$ & $[-0.0326, 0.0035]$ \\
Llama-3.1-8B & $-0.0146$ & $[-0.0368, 0.0085]$ \\
\bottomrule
\end{tabular}
\end{adjustbox}
\caption{Paired uncertainty for matched visual textualization. $\Delta_{\mathrm{V}}$ is the seed-ensemble RMSE difference between $Q{+}D_{\mathrm{VL7B}}$ and $Q$; negative values favor visual textualization.}
\label{tab:app-textualization-ci}
\end{table}

\begin{table}[t]
\centering
\small
\begin{adjustbox}{width=0.48\linewidth,center}
\begin{tabular}{@{}lrr@{}}
\toprule
VLM & $\Delta_{\mathrm{M}}\downarrow$ & 95\% CI \\
\midrule
Qwen2.5-VL-3B & $-0.0510$ & $[-0.0854,-0.0115]$ \\
Qwen2.5-VL-7B & $-0.0260$ & $[-0.0554, 0.0141]$ \\
Qwen3-VL-4B & $-0.0171$ & $[-0.0399, 0.0078]$ \\
Qwen3-VL-8B & $-0.0116$ & $[-0.0317, 0.0101]$ \\
PaliGemma2-3B & $-0.0266$ & $[-0.0557, 0.0029]$ \\
InternVL3-2B & $-0.0304$ & $[-0.0589,-0.0018]$ \\
InternVL2.5-4B & $-0.0341$ & $[-0.0639,-0.0055]$ \\
InternVL3-8B & $-0.0321$ & $[-0.0550,-0.0071]$ \\
InternVL3.5-4B & $-0.0293$ & $[-0.0557,-0.0023]$ \\
InternVL3.5-8B & $-0.0336$ & $[-0.0507,-0.0164]$ \\
\bottomrule
\end{tabular}
\end{adjustbox}
\caption{Paired uncertainty for image-native adaptation breadth. $\Delta_{\mathrm{M}}$ is the seed-ensemble RMSE difference between attention+MLP and attention-only LoRA; negative values favor broader adaptation.}
\label{tab:app-vlm-mlp-ci}
\end{table}

For completeness, exact one-sided sign tests give $p=0.031$ for the 5/5 direction of the matched $Q+D_{\mathrm{VL7B}}$ comparison and $p=0.001$ for the 10/10 direction of attention+MLP adaptation. These values are descriptive: systems share the same data, and several share model families, so they are not independent scientific replicates.

\subsection{Rasch-Label Reliability and Sensitivity}
\label{app:label-uncertainty}

The Rasch estimates include both the point estimate $\beta_i$ and its reported standard error. Across all 725 retained items, the response count has mean 1,879, median 2,155, and range 203--2,966. The difficulty standard error has mean 0.061, median 0.054, and range 0.043--0.162; its 90th percentile is 0.093. On the 145 test items, the corresponding mean, median, and range are 0.061, 0.054, and 0.043--0.161.

We conduct a sensitivity analysis using the seed-ensemble predictions of the three leading systems. We exclude the 10\% or 20\% of test items with the largest reported difficulty standard errors and recompute RMSE without retraining. We also correlate each item's reported standard error with its squared prediction error.

\tabref{tab:app-label-sensitivity} reports the recomputed RMSE values and the association between label standard error and squared prediction error.

\begin{table*}[t]
\centering
\small
\begin{adjustbox}{width=0.6\linewidth,center}
\begin{tabular}{lcccc}
\toprule
System & All 145 $\downarrow$ & Exclude top 10\% SE $\downarrow$ & Exclude top 20\% SE $\downarrow$ & $\rho(\mathrm{SE}, e_i^2)$ \\
\midrule
$Q$ & 0.5054 & 0.5209 & 0.5199 & 0.008 \\
$Q+D_{\mathrm{VL7B}}$ & 0.4975 & 0.5123 & 0.5150 & 0.111 \\
$I+Q$ & \textbf{0.4907} & \textbf{0.5055} & \textbf{0.5002} & 0.019 \\
\bottomrule
\end{tabular}
\end{adjustbox}
\caption{Sensitivity to reported Rasch difficulty standard error. Values use seed-ensemble predictions; they differ slightly from means of three run-level RMSEs. Removing high-SE items does not change the ordering, and standard error is only weakly associated with squared prediction error.}
\label{tab:app-label-sensitivity}
\end{table*}

Absolute RMSE increases after trimming because the high-SE items are not the items with the largest model errors; this does not indicate worse labels after trimming. The training-mean predictor obtains 0.8199 RMSE on the complete test set (the training-median predictor obtains 0.8220), providing a common constant baseline for the learned systems.

\section{Additional Results}
\label{app:detailed-results}

\subsection{Question-Text and Visual-Textualization Systems}
\label{app:text-results}

\tabref{tab:app-text-main} reports complete three-seed RMSE and Spearman results for the five matched text models under $Q$, $Q+D_{\mathrm{GPT}}$, and $Q+D_{\mathrm{VL7B}}$. It also includes the strongest $Q$-only attention+MLP control.

\begin{table*}[t]
\centering
\small
\begin{adjustbox}{width=0.94\linewidth,center}
\begin{tabular}{lcccccc}
\toprule
& \multicolumn{2}{c}{$Q$} & \multicolumn{2}{c}{$Q+D_{\mathrm{GPT}}$} & \multicolumn{2}{c}{$Q+D_{\mathrm{VL7B}}$} \\
\cmidrule(lr){2-3}\cmidrule(lr){4-5}\cmidrule(lr){6-7}
Text model & RMSE $\downarrow$ & Spearman $\uparrow$ & RMSE $\downarrow$ & Spearman $\uparrow$ & RMSE $\downarrow$ & Spearman $\uparrow$ \\
\midrule
Qwen2.5-3B & $0.5367\pm0.0132$ & $0.7513\pm0.0104$ & $0.5343\pm0.0106$ & $0.7559\pm0.0135$ & $0.5266\pm0.0051$ & $0.7553\pm0.0047$ \\
Llama-3.2-3B & $0.5615\pm0.0152$ & $0.7284\pm0.0189$ & $0.5608\pm0.0081$ & $0.7322\pm0.0152$ & $0.5335\pm0.0022$ & $0.7672\pm0.0027$ \\
Qwen2.5-7B & $0.5226\pm0.0067$ & $0.7696\pm0.0054$ & $0.5258\pm0.0057$ & $0.7655\pm0.0073$ & $0.5169\pm0.0112$ & $0.7791\pm0.0089$ \\
Qwen3-8B & $0.5380\pm0.0053$ & $0.7484\pm0.0048$ & $0.5165\pm0.0071$ & $0.7769\pm0.0081$ & $0.5210\pm0.0062$ & $0.7749\pm0.0013$ \\
Llama-3.1-8B & $0.5210\pm0.0065$ & $0.7654\pm0.0063$ & $0.5298\pm0.0055$ & $0.7653\pm0.0032$ & $\mathbf{0.5059\pm0.0075}$ & $\mathbf{0.7939\pm0.0084}$ \\
Llama-3.1-8B, +MLP $Q$ & $\mathbf{0.5167\pm0.0147}$ & $\mathbf{0.7727\pm0.0124}$ & --- & --- & --- & --- \\
\bottomrule
\end{tabular}
\end{adjustbox}
\caption{Complete three-seed text-model results. The first five rows hold attention-only LoRA fixed across inputs; the final row reports the strongest $Q$-only adaptation control summarized in \tabref{tab:route-comparison}. $Q+D$ is text-valued but vision-informed because each $D$ is generated from the original image.}
\label{tab:app-text-main}
\end{table*}

The attention+MLP text ablation changes Llama-3.1-8B $Q$ from $0.5210\pm0.0065$ to $0.5167\pm0.0147$ RMSE, with Spearman $0.7727\pm0.0124$. Its seed-ensemble difference is $-0.0066$ with a 95\% interval of $[-0.0314,0.0214]$.

\subsection{Raw-OCR Control}
\label{app:ocr-control}

The main $Q$ representation is deliberately stronger than uncorrected OCR. To quantify the difference, we apply EasyOCR to all original images without manual correction, replace $Q$ with the resulting text for both training and testing, and reuse the fixed Llama-3.1-8B attention-LoRA recipe.

\tabref{tab:app-ocr} compares the curated $Q$ representation with this raw-OCR input while holding the downstream model, split, and training recipe fixed.

\begin{table*}[t]
\centering
\small
\begin{adjustbox}{width=0.46\linewidth,center}
\begin{tabular}{lcc}
\toprule
Input to Llama-3.1-8B & RMSE $\downarrow$ & Spearman $\uparrow$ \\
\midrule
Curated transcription $Q$ & $\mathbf{0.5210\pm0.0065}$ & $\mathbf{0.7654\pm0.0063}$ \\
Uncorrected EasyOCR & $0.5898\pm0.0059$ & $0.7082\pm0.0041$ \\
\bottomrule
\end{tabular}
\end{adjustbox}
\caption{Matched transcription-quality control (three-seed mean $\pm$ SD). The downstream model, split, and training recipe are fixed.}
\label{tab:app-ocr}
\end{table*}

The raw-OCR control is not a fourth representation strategy: both conditions reduce the item to text, but they differ in transcription fidelity. It shows that the competitive $Q$ baseline cannot be reproduced by substituting an off-the-shelf OCR string.

\subsection{VLM Input-Route Baselines}
\label{app:image-input-results}

\tabref{tab:app-vlm-inputs} reports the complete attention-only VLM results for $I$, $I+Q$, and $I+Q+D$, allowing input route to be compared before broader adaptation is introduced.

\begin{table*}[t]
\centering
\scriptsize
\begin{adjustbox}{width=0.94\linewidth,center}
\begin{tabular}{lcccccc}
\toprule
& \multicolumn{2}{c}{$I$} & \multicolumn{2}{c}{$I+Q$} & \multicolumn{2}{c}{$I+Q+D$} \\
\cmidrule(lr){2-3}\cmidrule(lr){4-5}\cmidrule(lr){6-7}
VLM, attention-only & RMSE $\downarrow$ & Spearman $\uparrow$ & RMSE $\downarrow$ & Spearman $\uparrow$ & RMSE $\downarrow$ & Spearman $\uparrow$ \\
\midrule
Qwen2.5-VL-3B & $0.5759\pm.0058$ & $0.7113\pm.0036$ & $0.5762\pm.0072$ & $0.7117\pm.0093$ & $0.5751\pm.0106$ & $0.7067\pm.0148$ \\
Qwen2.5-VL-7B & $0.5210\pm.0030$ & $0.7667\pm.0063$ & $0.5233\pm.0077$ & $0.7649\pm.0049$ & $0.5273\pm.0069$ & $0.7614\pm.0043$ \\
Qwen3-VL-4B & $0.5630\pm.0050$ & $0.7197\pm.0054$ & $0.5461\pm.0089$ & $0.7406\pm.0093$ & $0.5482\pm.0050$ & $0.7392\pm.0044$ \\
Qwen3-VL-8B & $0.5512\pm.0011$ & $0.7398\pm.0014$ & $0.5379\pm.0076$ & $0.7564\pm.0088$ & $0.5306\pm.0035$ & $0.7619\pm.0100$ \\
PaliGemma2-3B & $0.5715\pm.0053$ & $0.7107\pm.0054$ & $0.5857\pm.0147$ & $0.6943\pm.0128$ & $0.5973\pm.0082$ & $0.6783\pm.0137$ \\
InternVL3-2B & $0.5718\pm.0147$ & $0.7069\pm.0157$ & $0.5765\pm.0161$ & $0.7038\pm.0155$ & $0.5796\pm.0115$ & $0.7011\pm.0106$ \\
InternVL2.5-4B & $0.5554\pm.0054$ & $0.7270\pm.0023$ & $0.5562\pm.0093$ & $0.7280\pm.0052$ & $0.5656\pm.0094$ & $0.7192\pm.0037$ \\
InternVL3-8B & $0.5805\pm.0083$ & $0.6956\pm.0150$ & $0.5806\pm.0112$ & $0.6988\pm.0189$ & $0.5830\pm.0047$ & $0.6969\pm.0083$ \\
InternVL3.5-4B & $0.5414\pm.0172$ & $0.7463\pm.0221$ & $0.5408\pm.0057$ & $0.7453\pm.0113$ & $0.5545\pm.0016$ & $0.7312\pm.0066$ \\
InternVL3.5-8B & $0.6275\pm.0291$ & $0.6375\pm.0345$ & $0.6354\pm.0087$ & $0.6362\pm.0251$ & $0.6290\pm.0089$ & $0.6395\pm.0130$ \\
\bottomrule
\end{tabular}
\end{adjustbox}
\caption{Attention-only input-route baselines (three-seed mean $\pm$ SD). Small InternVL models use the 10-epoch recipe selected from training-side learning curves.}
\label{tab:app-vlm-inputs}
\end{table*}

Adding $Q$ to $I$ does not yield a uniform change across the displayed attention-only means. This result motivates treating visual access and successful visual adaptation as distinct questions.

\subsection{Broad Language-Side Adaptation}
\label{app:vlm-mlp-full}

\tabref{tab:app-vlm-mlp-full} reports complete attention+MLP results for all ten VLMs on the canonical $I+Q$ image-native input. These are the broad-adaptation values summarized in \tabref{tab:vlm-mlp}.

\begin{table*}[t]
\centering
\small
\begin{adjustbox}{width=0.5\linewidth,center}
\begin{tabular}{lcc}
\toprule
VLM, $I+Q$, attention+MLP & RMSE $\downarrow$ & Spearman $\uparrow$ \\
\midrule
Qwen2.5-VL-3B & $0.5280\pm0.0105$ & $0.7586\pm0.0073$ \\
Qwen2.5-VL-7B & $\mathbf{0.4966\pm0.0062}$ & $\mathbf{0.7853\pm0.0079}$ \\
Qwen3-VL-4B & $0.5306\pm0.0011$ & $0.7547\pm0.0058$ \\
Qwen3-VL-8B & $0.5277\pm0.0036$ & $0.7631\pm0.0049$ \\
PaliGemma2-3B & $0.5611\pm0.0093$ & $0.7206\pm0.0125$ \\
InternVL3-2B & $0.5452\pm0.0101$ & $0.7384\pm0.0109$ \\
InternVL2.5-4B & $0.5214\pm0.0065$ & $0.7590\pm0.0088$ \\
InternVL3-8B & $0.5500\pm0.0162$ & $0.7268\pm0.0169$ \\
InternVL3.5-4B & $0.5111\pm0.0107$ & $0.7685\pm0.0095$ \\
InternVL3.5-8B & $0.6137\pm0.0258$ & $0.6513\pm0.0369$ \\
\bottomrule
\end{tabular}
\end{adjustbox}
\caption{Complete attention+MLP results (three-seed mean $\pm$ SD). These are the broad-adaptation values summarized in \tabref{tab:vlm-mlp}.}
\label{tab:app-vlm-mlp-full}
\end{table*}

\subsection{Pooling and Vision-Scope Ablations}
\label{app:vlm-ablation-results}

To further diagnose the gains from broader language-side adaptation, \tabref{tab:interface-ablation} compares attention+MLP LoRA with two alternative changes on representative 7--8B VLMs. Text-token pooling excludes image placeholder tokens when averaging the final hidden states and lowers the point-estimate RMSE for both models. Extending LoRA to the vision encoder provides no consistent benefit. The best variant differs across the two VLMs: attention+MLP adaptation performs best for Qwen2.5-VL-7B, whereas text-token pooling performs best for Qwen3-VL-8B. None of the six paired intervals excludes zero. These results point to language-side adaptation and readout construction, rather than insufficient vision-encoder adaptation, as the more promising sources of improvement, although this limited comparison does not identify a single bottleneck.

\begin{table}[H]
\centering
\scriptsize
\begin{adjustbox}{width=0.78\linewidth,center}
\begin{tabular}{llccc}
\toprule
Model & Adaptation & RMSE $\downarrow$ & $\Delta$ $\downarrow$ & 95\% CI \\
\midrule
Qwen2.5-VL-7B & Attention only & 0.5233 & --- & --- \\
 & Attention + MLP & \textbf{0.4966} & $-0.0260$ & $[-0.0554, 0.0141]$ \\
 & Text-token pooling & 0.5085 & $-0.0165$ & $[-0.0389, 0.0060]$ \\
 & Attention + vision LoRA & 0.5222 & $-0.0031$ & $[-0.0131, 0.0079]$ \\
\midrule
Qwen3-VL-8B & Attention only & 0.5379 & --- & --- \\
 & Attention + MLP & 0.5277 & $-0.0116$ & $[-0.0317, 0.0101]$ \\
 & Text-token pooling & \textbf{0.5216} & $-0.0190$ & $[-0.0400, 0.0022]$ \\
 & Attention + vision LoRA & 0.5475 & $+0.0079$ & $[-0.0052, 0.0210]$ \\
\bottomrule
\end{tabular}
\end{adjustbox}
\caption{Adaptation, pooling, and vision-scope ablations for representative VLMs on $I+Q$. RMSE values are three-seed means; $\Delta$ values and paired confidence intervals use seed-ensemble predictions relative to attention-only LoRA.}
\label{tab:interface-ablation}
\end{table}

\subsection{Test-Time Image Interventions}
\label{app:image-intervention}

For each trained Qwen2.5-VL-7B seed, we hold the adapter and regression head fixed and alter only the held-out images. The \emph{blank} condition substitutes a $448\times448$ white canvas. The \emph{shuffle} condition applies one fixed random derangement to the 145 test images, so that every item receives another item's full question image. Text inputs are unchanged. We report variation across training seeds but do not estimate variation across alternative derangements.

\tabref{tab:app-image-intervention} reports original, blank, and shuffled-image results for attention-only $I+Q$, attention+MLP $I+Q$, and attention-only image-only prediction.

\begin{table*}[t]
\centering
\small
\begin{adjustbox}{width=0.64\linewidth,center}
\begin{tabular}{llcc}
\toprule
Trained input & Test image & RMSE $\downarrow$ & Spearman $\uparrow$ \\
\midrule
$I+Q$, attention & Original & $0.5233\pm0.0077$ & $0.7649\pm0.0049$ \\
 & Blank & $1.0746\pm0.1299$ & $0.6492\pm0.0080$ \\
 & Shuffled & $0.9552\pm0.0044$ & $0.1121\pm0.0093$ \\
\midrule
$I+Q$, attention+MLP & Original & $\mathbf{0.4966\pm0.0062}$ & $\mathbf{0.7853\pm0.0079}$ \\
 & Blank & $0.9447\pm0.0475$ & $0.6788\pm0.0069$ \\
 & Shuffled & $0.9643\pm0.0059$ & $0.1118\pm0.0065$ \\
\midrule
$I$, attention & Original & $0.5210\pm0.0030$ & $0.7667\pm0.0063$ \\
 & Blank & $1.2629\pm0.2093$ & undefined \\
 & Shuffled & $1.0528\pm0.0092$ & $-0.0027\pm0.0148$ \\
\bottomrule
\end{tabular}
\end{adjustbox}
\caption{Test-time image interventions (three-seed mean $\pm$ SD). Blank-image predictions are constant within each image-only seed, so their Spearman correlation is undefined. These deliberately unnatural inputs test image dependence rather than estimate the benefit of vision under the data distribution.}
\label{tab:app-image-intervention}
\end{table*}

\subsection{Items With and Without Identified Visual Components}
\label{app:subgroups}

\tabref{tab:app-subgroups} compares seed-ensemble RMSE on the 76 items with an identified visual component and the 69 remaining test items. The comparison is diagnostic rather than a routing rule because the direction varies across models and adaptation choices.

\begin{table*}[t]
\centering
\small
\begin{adjustbox}{width=0.62\linewidth,center}
\begin{tabular}{lcc}
\toprule
System & Visual (76) $\downarrow$ & No identified visual (69) $\downarrow$ \\
\midrule
Llama-3.1-8B, $Q+D_{\mathrm{VL7B}}$ & \textbf{0.4942} & 0.5011 \\
Qwen2.5-VL-7B, $I+Q$, text pooling & 0.5035 & \textbf{0.4964} \\
Qwen2.5-VL-7B, $I+Q$, attention+MLP & 0.5266 & \textbf{0.4478} \\
InternVL2.5-4B, $I+Q$, attention+MLP & \textbf{0.4979} & 0.5366 \\
InternVL3.5-8B, $I+Q$, attention+MLP & \textbf{0.5438} & 0.6505 \\
\bottomrule
\end{tabular}
\end{adjustbox}
\caption{Seed-ensemble RMSE by coarse item group. Bold marks the lower RMSE within each row; the direction varies across model families and adaptation choices.}
\label{tab:app-subgroups}
\end{table*}

\subsection{Difficulty Strata and Representation Complementarity}
\label{app:complementarity}

We divide the 145 held-out items into five equal-sized groups based on their gold Rasch difficulty. \tabref{tab:app-difficulty-strata} shows a different ordering across the target range: image-native prediction has lower RMSE in the two extreme groups, whereas visual textualization has lower RMSE in the middle three. Each group contains 29 items, so the analysis is intended to characterize errors across the difficulty range rather than define a routing rule.

\begin{table}[H]
\centering
\small
\begin{adjustbox}{width=0.56\linewidth,center}
\begin{tabular}{lccc}
\toprule
Gold difficulty range ($n=29$) & $Q\downarrow$ & $Q+D_{\mathrm{VL7B}}\downarrow$ & $I+Q\downarrow$ \\
\midrule
$(-2.19,-0.86]$ & 0.6801 & 0.6832 & \textbf{0.5896} \\
$(-0.86,-0.36]$ & 0.3326 & \textbf{0.3108} & 0.4146 \\
$(-0.36,0.05]$ & 0.4404 & \textbf{0.3706} & 0.3994 \\
$(0.05,0.50]$ & 0.4462 & \textbf{0.4283} & 0.4840 \\
$(0.50,1.38]$ & 0.5577 & 0.5943 & \textbf{0.5389} \\
\bottomrule
\end{tabular}
\end{adjustbox}
\caption{Seed-ensemble RMSE by gold-difficulty quintile. The $Q$ and $I+Q$ columns use their strongest attention+MLP systems; $Q+D_{\mathrm{VL7B}}$ uses its strongest attention-only system. Bold marks the lowest RMSE in each row.}
\label{tab:app-difficulty-strata}
\end{table}

The two leading visual representations also leave complementary residual errors. Image-native modeling has lower absolute error on 74 of 145 items and visual textualization on 71. Across their three matched seeds, the preference is unanimous for 37 image-native wins and 40 textualization wins; the other 68 items change winner across seeds. Their fixed equal-weight average requires no fitted fusion parameters and reaches 0.4780 RMSE, but its paired intervals relative to either component cross zero. We therefore use the average as a complementarity diagnostic rather than a separately trained fusion system. An oracle that selects the lower-error prediction for each item uses the test target and is unattainable; it is reported only to quantify headroom.

\tabref{tab:app-complementarity} reports the two component systems, their fixed average, and oracle diagnostics that quantify unattainable item-level headroom.

\begin{table}[H]
\centering
\small
\begin{adjustbox}{width=0.5\linewidth,center}
\begin{tabular}{lc}
\toprule
Seed-ensemble diagnostic & RMSE $\downarrow$ \\
\midrule
Visual textualization, $Q+D_{\mathrm{VL7B}}$ & 0.4975 \\
Image-native, $I+Q$ & 0.4907 \\
Fixed average of the two & \textbf{0.4780} \\
Two-system per-item oracle & 0.4090 \\
Three-system oracle (also including $Q$) & 0.3908 \\
\bottomrule
\end{tabular}
\end{adjustbox}
\caption{Complementarity diagnostics. The fixed average differs from visual-textualization and image-native prediction by $-0.0195$ RMSE (95\% CI $[-0.0427,0.0031]$) and $-0.0127$ ($[-0.0346,0.0102]$), respectively. Oracle rows use gold labels and are not prediction systems.}
\label{tab:app-complementarity}
\end{table}

Description verbosity does not explain when textualization helps. Among the 76 test items with an identified visual component, the Spearman correlation between the actual $D_{\mathrm{VL7B}}$ character length and the per-item squared-error reduction over $Q$ is $0.121$ ($p=0.30$). Short, middle, and long length tertiles have RMSE changes of $+0.008$, $-0.030$, and $+0.006$, respectively, providing no monotonic length--benefit relation.

\subsection{Deterministic Taxonomy and Error Cases}
\label{app:error-cases}

For the systematic analysis in \tabref{tab:error-taxonomy}, we assign mutually exclusive categories from the verified item representations in a fixed order. An item is a \emph{visual answer options} item when an answer choice is image-valued or the description explicitly identifies visual options. The next category covers plots, tables, coordinate axes, number lines, and grids. The geometry category covers named shapes and geometric relations such as angles, parallel or perpendicular lines, radii, and vertices. Remaining items with an identified visual component are \emph{other visual}; all others have \emph{no identified visual component}. Category assignment does not use difficulty labels or model predictions, but the assignments have not been independently human-audited.

\begin{table}[H]
\centering
\small
\begin{adjustbox}{width=0.72\linewidth,center}
\begin{tabular}{lrrrr}
\toprule
Item type & $n$ & $Q\downarrow$ & $Q{+}D_{\mathrm{VL7B}}\downarrow$ & $I{+}Q\downarrow$ \\
\midrule
Visual answer options & 4 & \textbf{0.2482} & 0.3090 & 0.3696 \\
Plots/tables/number lines & 34 & 0.5644 & \textbf{0.5452} & 0.5769 \\
Geometry diagrams & 26 & 0.5087 & \textbf{0.5053} & 0.5292 \\
Other visual & 12 & 0.3819 & \textbf{0.3453} & 0.4008 \\
\midrule
No identified visual & 69 & 0.5034 & 0.5011 & \textbf{0.4478} \\
\bottomrule
\end{tabular}
\end{adjustbox}
\caption{Error analysis by deterministic item type. Values are seed-ensemble RMSE for the three leading systems; bold marks the lowest value within each row. Small subgroups, especially visual answer options, should not be interpreted as stable rankings.}
\label{tab:error-taxonomy}
\end{table}

Across all 76 explicitly visual items, visual textualization improves over $Q$ from 0.5073 to 0.4942 RMSE ($\Delta=-0.0131$, 95\% CI $[-0.0598,0.0344]$), while image-native prediction reaches 0.5266. Visual textualization has the lowest point estimate for the three adequately sized explicit-visual categories, but none of their individual intervals establishes a precise category-level ordering.

Conversely, the image-native system's largest aggregate gain occurs on the 69 items without an identified visual component, from 0.5034 to 0.4478 RMSE ($\Delta=-0.0555$, 95\% CI $[-0.1189,0.0037]$). These items can still contain rendered equations, box symbols, spatial layout, and visually redundant question text. The pattern therefore argues against attributing the image-native point estimate only to explicit diagrams, but remains suggestive because the categories are automatically derived and the interval includes zero.

Inspection of the largest per-item error reductions suggests different mechanisms. For item 83, $D_{\mathrm{VL7B}}$ makes the reflex-angle interpretation explicit and reduces absolute error by 0.39. For items 132 and 246, image-native prediction better captures a grid-based area relation and equality tick marks, reducing absolute error by 0.45 and 0.44. \tabref{tab:app-error-cases} reports these and additional cases.

\begin{table}[H]
\centering
\small
\begin{adjustbox}{width=0.62\linewidth,center}
\begin{tabular}{rlrrrr}
\toprule
Item & Type/cue & Gold & $Q$ & $Q{+}D_{\mathrm{VL7B}}$ & $I{+}Q$ \\
\midrule
83 & Reflex angle & $-0.140$ & 0.279 & $\mathbf{-0.167}$ & 0.376 \\
132 & Area on grid & 0.590 & $-0.239$ & 0.018 & $\mathbf{0.210}$ \\
246 & Equality tick marks & 0.908 & $-0.020$ & 0.373 & $\mathbf{0.417}$ \\
668 & Angles on grid & $-1.084$ & 0.625 & $\mathbf{0.272}$ & 0.654 \\
3 & Box notation, no figure & $-1.735$ & $-0.764$ & $-0.742$ & $\mathbf{-1.332}$ \\
\bottomrule
\end{tabular}
\end{adjustbox}
\caption{Qualitative cases selected from the largest reductions in absolute error relative to the strongest $Q$ system while spanning several taxonomy groups. Bold marks the closest prediction to the Rasch target.}
\label{tab:app-error-cases}
\end{table}

\paragraph{All visual-answer-option cases.}
The four test items in this category are unusual because the answer choices cannot be transcribed as ordinary text. The curated $Q$ therefore includes a short description of each option---coordinates for item 62, dimensions and perpendicular heights for items 215 and 869, and colored-grid configurations for item 227. This makes $Q$ a comparatively rich reduced interface and helps explain why adding another description is redundant on three of four items. The complete group is shown in \tabref{tab:app-visual-options}; with $n=4$, its aggregate ordering is not a stable estimate of a population effect.

\begin{table}[H]
\centering
\small
\begin{adjustbox}{width=0.6\linewidth,center}
\begin{tabular}{rlrrrr}
\toprule
Item & Visual task & Gold & $Q$ & $Q{+}D_{\mathrm{VL7B}}$ & $I{+}Q$ \\
\midrule
62 & Rotation on grid & 0.222 & \textbf{0.136} & 0.062 & 0.390 \\
215 & Triangle areas & 0.511 & \textbf{0.347} & 0.342 & 0.770 \\
227 & Reflection symmetry & $-0.836$ & $\mathbf{-1.284}$ & $-1.407$ & $-1.466$ \\
869 & Triangle areas & 0.545 & 0.440 & \textbf{0.499} & 0.780 \\
\bottomrule
\end{tabular}
\end{adjustbox}
\caption{All four visual-answer-option test items. Bold marks the prediction with the smallest absolute error relative to the Rasch target, not the numerically smallest prediction.}
\label{tab:app-visual-options}
\end{table}

\subsection{Auxiliary Output and Fusion Baselines}
\label{app:auxiliary-results}

Generic visual features and frozen multimodal representations contain difficulty signal but remain behind task-adapted systems. The best frozen generic vision encoder obtains 0.557 RMSE, and the best frozen VLM plus external regressor obtains 0.543. Concatenating frozen text, vision, and VLM representations reaches 0.552. Fine-tuning a VLM to generate a standardized numeric string reaches 0.601 at best. These experiments use the same item split but serve as auxiliary paradigm checks rather than entries in the unified three-seed comparison.

\tabref{tab:app-auxiliary} summarizes the strongest representative from each auxiliary paradigm and identifies the input and model used by that representative.

\begin{table}[H]
\centering
\small
\begin{adjustbox}{width=0.52\linewidth,center}
\begin{tabular}{llcc}
\toprule
Paradigm & Best representative & RMSE $\downarrow$ & Spearman $\uparrow$ \\
\midrule
Text encoder & DeBERTa-v3-base, $Q+D_{\mathrm{VL7B}}$ & 0.541 & 0.753 \\
Frozen text LLM & Llama-3.2-3B, $Q+D_{\mathrm{VL7B}}$ & 0.529 & 0.783 \\
Generic vision encoder & DINOv2-large, $I$ & 0.557 & 0.721 \\
Frozen VLM & Qwen2.5-VL-3B, $I+Q+D$ & 0.543 & 0.747 \\
Scalar generation & Qwen3-VL-8B, $I+Q$ & 0.601 & 0.662 \\
Late fusion & DINOv2 + Qwen2.5-VL-3B & 0.552 & 0.745 \\
\bottomrule
\end{tabular}
\end{adjustbox}
\caption{Best auxiliary representative in each modeling paradigm. These runs use the same held-out item split but predate the unified three-seed protocol and are not used for its paired confidence intervals.}
\label{tab:app-auxiliary}
\end{table}

\end{document}